\documentclass{article}

\usepackage{spconf}
\usepackage{cite}
\usepackage{graphicx}
\usepackage{amsmath}
\usepackage{amssymb}
\usepackage{booktabs}
\usepackage{multirow}
\usepackage{multicol}
\usepackage{comment}
\usepackage{subfig}
\usepackage[table]{xcolor}
\usepackage[linesnumbered,ruled,vlined]{algorithm2e}
\usepackage[pagebackref,breaklinks,colorlinks]{hyperref}

\title{SWIS: Self-Supervised Representation Learning for Writer Independent Offline Signature Verification}
\name{Siladittya Manna$^{\dagger}$ \qquad Soumitri Chattopadhyay$^{\star}$ \qquad Saumik Bhattacharya$^{\diamond}$ \qquad Umapada Pal$^{\dagger}$}
  
  \address{$^{\dagger}$ Indian Statistical Institute, Kolkata; $^{\diamond}$Indian Institute of Technology, Kharagpur; $^{\star}$Jadavpur University}
\begin{document}
%\ninept
%
\maketitle

% As a general rule, do not put math, special symbols or citations
% in the abstract
\begin{abstract}
Writer independent offline signature verification is one of the most challenging tasks in pattern recognition as there is often a scarcity of training data. To handle such data scarcity problem,  in this paper, we propose a novel self-supervised  learning (SSL) framework for writer independent offline signature verification.  To our knowledge, this is  the first attempt to utilize self-supervised setting for the signature verification task. The objective of self-supervised representation learning from the signature images is achieved by minimizing the cross-covariance between two random variables belonging to different feature directions and ensuring a positive cross-covariance between the random variables denoting the same feature direction. This ensures that the features are decorrelated linearly and the redundant information is discarded. Through experimental results on different data sets, we obtained encouraging results. 

\textbf{\textit{Keywords}} - Self-supervised, Cross-covariance, Decorrelation, Writer-independent, SVM
\end{abstract}

 %% objective : 1. training time is less, 2. advtange of ssl in view of data

% no keywords

% For peer review papers, you can put extra information on the cover
% page as needed:
% \ifCLASSOPTIONpeerreview
% \begin{center} \bfseries EDICS Category: 3-BBND \end{center}
% \fi
%
% For peerreview papers, this IEEEtran command inserts a page break and
% creates the second title. It will be ignored for other modes.

\section{Introduction}
Signature verification has been used as one of the most essential steps for identity verification of person-specific documents like forms, bank cheques, or even the individual themselves. This makes signature verification an important task in domain of computer vision and pattern recognition. There are mainly two types of signature verification processes: (1) offline and (2) online. In offline signature verification, the input is basically a 2D image which is scanned from the original signature or captured into an image by some electronic device. Whereas, in online signature verification, the writer usually pens down his signature on an electronic tablet using a stylus and the information is recorded at some regular timestep along with the position of the stylus. 

%Due to the limited amount of available information, offline signature verification is typically more challenging than online verification process. In offline signature verification, only an image is provided, which may be the case in many real-life scenarios. Furthermore, in offline scheme, only original signatures are available as references initially and that too in small numbers. It is generally not possible to obtain forged signatures beforehand and use it as reference. 

Offline signature verification can again be divided into two types: (1) Writer dependent and (2) writer independent.  In writer dependent scenario, the system needs to be updated and retrained for every new user signature that gets added to the system. This makes the process cumbersome and less feasible. However, in writer independent scenario, a generalized system needs to be built which can differentiate between genuine and forged signatures without repeated retraining.

% \textit{\textbf{LITERATURE REVIEW}}
Most researchers have leveraged supervised learning methods \cite{rantzsch2016signature, dey2017signet, ruiz2020off, wan2021learning, parcham2021cbcapsnet, bhunia2019signature} for offline signature verification. While handcrafted feature analyses have comprised the bulk of studies in this domain \cite{bhunia2019signature, alaei2017efficient, hafemann2017review, banerjee2021new}, various deep learning-based methods have also been proposed, particularly dwelling on metric learning approaches \cite{rantzsch2016signature, dey2017signet, ruiz2020off, wan2021learning}. Nevertheless, all the aforementioned works are fully supervised methods and therefore, share the common bottleneck of data scarcity. To this end, we demonstrate the first use of self-supervision for offline signature verification.

Self-supervised learning aims at developing a pre-training paradigm to learn a robust representation from an unlabelled corpus for generalization to any given downstream task. Widely studied in recent years, several pretext tasks have been proposed, such as solving jigsaw puzzles \cite{noroozi2016unsupervised}, image colorization \cite{zhang2016colorful} to name a few. Contrastive learning based self-supervised algorithms, like SimCLR \cite{chen2020simple}, MoCo \cite{he2020momentum} has also gained popularity, which aim at learning similarity between augmented views of the same image while distancing views from different images. \cite{zbontar2021barlow} aimed at simultaneously maximizing similarity and minimizing redundancy between embeddings of two distorted views of an image.  

In this work, we propose a self-supervised learning algorithm for offline writer-independent signature verification. Self-supervised learning is a sub-domain of unsupervised learning that aims at learning representations from the data without any ground truth or human annotations. %In particular, we aim to use a learning framework which helps the encoder to learn representations of signatures in terms of linearly decorrelated factors or dimensions in the feature space. Intuitively, this helps the encoder to represent a signature in terms of its several generative factors such that the redundant information content of each factor is minimized.  
As a skilled forgery is supposed to be very close to the genuine signature, it is necessary to distinguish between each constituting element of the signatures for correct classification. However, since it is not possible to obtain a large number of annotated genuine signatures from the individuals for training a large model, we use self-supervised learning for training the model to learn representations which are generalized for signatures over a large number of individuals. This work is the first of its kind to apply self-supervised learning framework for learning representations from signature images. Also, in the downstream stage, we do not use any siamese type architecture in the downstream task for the offline signature verification, and show the capability of the pretrained encoder to effectively cluster the genuine signatures of the different unknown writers.

The main contributions of this work are as follows: \vspace{-2.5mm}

\begin{itemize}
    \item A novel self-supervised approach is introduced here for offline writer independent signature verification purpose. \vspace{-2mm} 
    %\item We show that the proposed approach improves performance over the state-of-the art self-supervised contrastive learning approaches on the same task.
    \item To the best of our knowledge, this is the first work of the use of self-supervised learning in  signature verification. \vspace{-6mm}
      
     \item We have shown that the proposed SSL is  better than  the  state-of-the art self-supervised contrastive learning approaches used in Computer vision and Medical image analysis areas. \vspace{-2mm}
\end{itemize}

The rest of the paper is organized as follows. %Sec. \ref{sec:relworks} describes some of the works in literature on offline signature verification and self-supervised learning. 
Sec. \ref{sec:method} describes the self-supervised learning methodology that is used in this work. Sec. \ref{sec:exptdet} presents the details about the datasets we use. In Sec. \ref{sec:results}, we present the experimental results and the comparison with the base models. Finally, we conclude the paper in Sec. \ref{sec:concl}.

%\section{Related Works}
%\label{sec:relworks}

%\subsection{Offline Signature Verification}

\begin{comment}
Dey et al. \cite{dey2017signet} introduced a contrastive loss based convolutional Siamese network for handwritten signature verification. Ruiz et al. \cite{ruiz2020off} combined synthetic signature generation with siamese networks for the verification task. The authors of \cite{shariatmadari2019patch} proposed a hierarchical CNN to learn features from patches of genuine signatures. Zhu et al. \cite{zhu2020point} sought to tackle intra-writer variations by introducing a point-to-set metric for offline signature verification. Zois et al. \cite{zois2017parsimonious} and Berkay et al. \cite{berkay2018hybrid} explored sparse dictionary learning and hybrid two-channel CNNs respectively for signature verification, whereas the authors of \cite{wan2021learning} proposed two triplet losses, each to tackle random and skilled forgeries respectively. Other works include an interval symbolic representation and fuzzy similarity measure \cite{alaei2017efficient} based handcrafted feature engineering method; region-based metric learning \cite{liu2021offline}; a neuromotor equivariance inspired model \cite{diaz2016approaching}; a recurrent neural network architecture \cite{ghosh2021recurrent} and a graph neural network based approach \cite{roy2021offline}. 
\end{comment}

%\subsection{Self-supervised Learning}

\section{Methodology}
\label{sec:method}
In this section, we discuss the pre-processing and the algorithm steps that are used to train the proposed encoder. 

\subsection{Pretraining Methodology}
\label{subsec:pretr}

In signature images, it is essential to capture the stroke information from the different authors as well as to learn the variations in the signatures of the same individual. %This allows the model to learn representations which not only discriminates one author from another, but also helps differentiate between genuine and forged signatures of an individual. %To satisfy this objective, we resort to self-supervised contrastive learning algorithm SimCLR [] because of the simplcity of this algorithm. 
To feed the stroke information without any human supervision, we divided the signature images into patches of dimensions $32 \times 32$ with an overlap of 16 pixels from a signature image reshaped to $224 \times 224$. This gives 169 patches from a single image of dimensions $32 \times 32$. As the base encoder we choose ResNet-18 \cite{he2016deep}. When the patches are passed through the encoder, we obtain an output of $1 \times 1 \times 512$ from each patch. We rearrange the patches into a grid of $13 \times 13$ to obtain an output of shape $13 \times 13 \times 512$. After applying global average pooling (GAP), we obtain an output feature vector of dimension $1 \times 512$. This feature vector is then passed through a non-linear projector with 1 hidden layer and output dimension $512$ to obtain the final output. 

%The main objective behind dividing the signature image into overlapping patches is to extract stroke information without human annotations of the same. 

%To facilitate stroke information learning, we intend to decorrelate the output dimensions of the encoder such that each one gives linearly uncorrelated information. The encoder discards redundant features by minimizing the cross-covariance between each dimension and extracts only meaningful information from the input by creating a bottleneck of information flow from the input space to the feature space. %We normalize the feature vectors such that they lie within an unit hypersphere $\mathcal{S}^{D}$, where $D$ is the dimension of the feature vector. 

For forming positive pairs, we augment a single signature image in two randomly chosen augmentations. The augmentation details are mentioned in Sec. \ref{subsec:preexptconfig}. The images are then divided into patches as mentioned before and then passed through the encoder and the projector. 

Thus, the proposed loss function has the form:

\begin{equation}
\label{eqn:lc}
\begin{split}
    \centering
    \mathcal{L}_{C} &= \frac{1}{N}\sum_{i=1}^{D} \left( \sum_{\substack{j=1\\j \neq i}}^{D} \left (\sum_{k=1}^{N} z_{k}^{i} \cdot {z'}_{k}^{j} \right)^2 +  \left ( \sum_{k=1}^{N} z_{k}^{i} \cdot {z'}_{k}^{i} - 1\right)^2 \right)\\
    %&=\frac{1}{N}\sum_{i=1}^{D} \left( \sum_{\substack{j=1\\j \neq i}}^{D} \left (\sum_{k=1}^{N} z_{k}^{i} \cdot {z'}_{k}^{j} \right)^2\right) \\ &+  \frac{1}{N}\sum_{i=1}^{D} \left( \left ( \sum_{k=1}^{N} z_{k}^{i} \cdot {z'}_{k}^{i} - 1\right)^2 \right)
\end{split}
\end{equation}

where $z_k^i$ is a scalar value at $i$-th dimension of the $k$-th centered and normalized feature vector $z_k$. Thus, the pre-processing steps before feeding the feature vector $z_k^i$ to the loss function are as follows
\begin{equation}
\centering
\begin{split}
    \overline{z}^i_k &= \frac{\widetilde{z}^i_k}{\sqrt{\sum_{k=1}^{N} (\widetilde{z}^i_k)^2}}\;\;\;\;\; \forall i \in [1,D]\\
    z^i_k &= \overline{z}^i_k - \mu_{z_k},\; \text{where} \; \mu_{z_k} = \frac{1}{N}\sum_{k=1}^{N} \overline{z}^i_k\;\;\; \forall i \in [1,D]\\
\end{split}
\label{eqn:norms}
\end{equation}

It is to be noted that $z^i_k$ and ${z'}^i_k$ are obtained from the each element of a positive pair. Thus, the proposed loss function does not optimize the terms of a cross-covariance matrix in the true meaning of the term. We can refer to this matrix as a Pseudo cross-covariance matrix.
%, and in \autoref{fig:pseudoccm}, we illustrate how this matrix is formed using the feature vectors. The diagonal elements (coloured white) constitute the second term of the loss function $\mathcal{L}_C$, whereas the rest of the non-diagonal elements constitute the first one. 

% \begin{figure}
%     \centering
%     \includegraphics[width = 0.8 \linewidth]{icpr2022_lossfn.png}
%     \caption{Illustration of Pseudo Cross-Covariance Matrix. The same coloured blocks indicate outputs from elements of a positive pair with the same parent image. One element from each positive pair taken together form $z$ and $z'$.}
%     \label{fig:pseudoccm}
% \end{figure}

From eq. \ref{eqn:lc}, we can see that optimizing the proposed loss function allows us to decorrelate the dimensions of the output. % by converting the pseudo cross-covariance matrix into a diagonal matrix with positive diagonal values. 
We treat each dimension as a random variable $Z_i$. As $Z_i$ is the output feature vector from the last Batch Normalization layer in the projecto, $Z_i \sim \mathcal{N}(0, 1)$. Normalizing $Z_i$ and subtracting mean along each dimension in Eqn. \ref{eqn:norms}, bring the feature vectors inside an unit hyper-sphere $\mathcal{S}^{D}$, where $D$ is the dimension of the feature vector, and centers each dimension at 0, i.e., $Z_i \sim \mathcal{N}(0, \sigma_i^2)$.  Since, we are making the cross-covariance matrix to an Indentity matrix,
\begin{equation}
\centering
\label{eqn:coveqn}
Cov(Z_i, Z_j) = 0 \Rightarrow \rho = 0
\end{equation}

\noindent
For Normal Random Variables $Z_i$,
\begin{equation}
\centering
\label{eqn:Eeqn}
\mathbb{E}[Z_i, Z_j] = \mathbb{E}[Z_i].\mathbb{E}[Z_j] \; \forall i,j \in [1,D] \wedge i \neq j
\end{equation}

\noindent
The diagonal terms of the cross-covariance matrix are optimised such that it equates to 1. Hence, the PDF of the feature vectors $f_{Z_1,..,Z_D} \sim \mathcal{N}(0, \mathcal{I}_{D \times D})$. Consequently, each output dimension becomes independent.

%This allows for positive correlation between the output in the same dimension for the samples in the positive pairs. %It also linearly decorrelates one dimension from another such that the representations learnt by the encoder discard redundant information and maximize the useful information. In other words, we can describe the representation learning process as learning linearly decorrelated generative factors of the input. 
%One thing to note is that, the cross-covariance and auto-covariance terms are not used in their true sense as the two samples used in a positive pair are not identical.

\subsection{Pretraining Model Architecture}
\label{subsec:premodarch}

\begin{figure}[tbp]
    \centering
    \includegraphics[width = \linewidth]{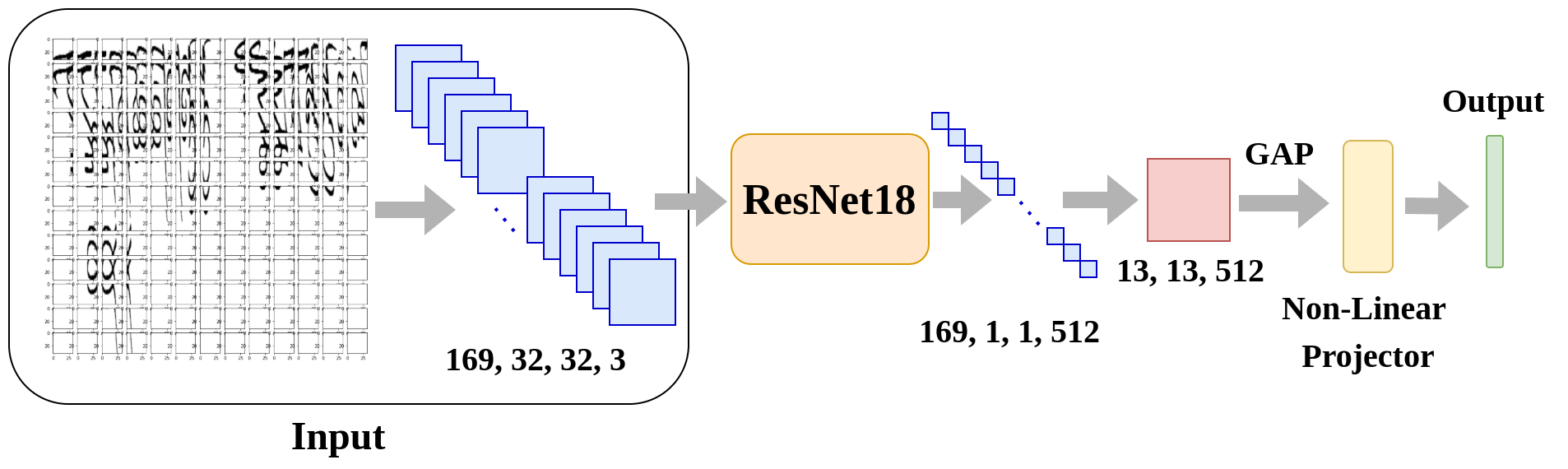}
    \caption{Model architecture used in the pretraining phase of the proposed method.}
    \label{fig:premodelarch}
\end{figure}

The model architecture used in the pretraining phase is given in \autoref{fig:premodelarch}. The diagram shows the input that is fed to the ResNet18 \cite{he2016deep} encoder. The input is reshaped to $169 \times 32 \times 32 \times 3$ before passing it through the encoder. \autoref{fig:premodelarch} also shows an example of the input used in the pretraining phase.

\subsection{Downstream Evaluation}
\label{subsec:downmethod}
For predicting whether a signature is forged or genuine, we take 8 reference signature for each user and use them to train a Support Vector Machine (SVM) classifier with radial basis function kernel. We assume that the user for which the signature is being verified is known. We also assume that the forged signature will be mapped outside the decision boundary of that particular user. If the user is predicted correctly and the signature is genuine, we count it as a correct prediction. Similarly, if the predicted user is not correct and the signature is actually forged, then also it is counted as a correct prediction. In all the other cases, the prediction is considered as wrong. 

By using a SVM classifier, we depend on the feature extraction capability of the pretrained encoder to express the input in terms of its linearly decorrelated factors. Whereas all the contemporary state-of-the-art supervised algorithms use siamese type architecture or supervised contrastive learning framework for the offline signature verification task. %Our approach also reduces the downstream inference time by a several orders of magnitude.
\begin{figure}[t]
    \centering
    \includegraphics[width = \linewidth]{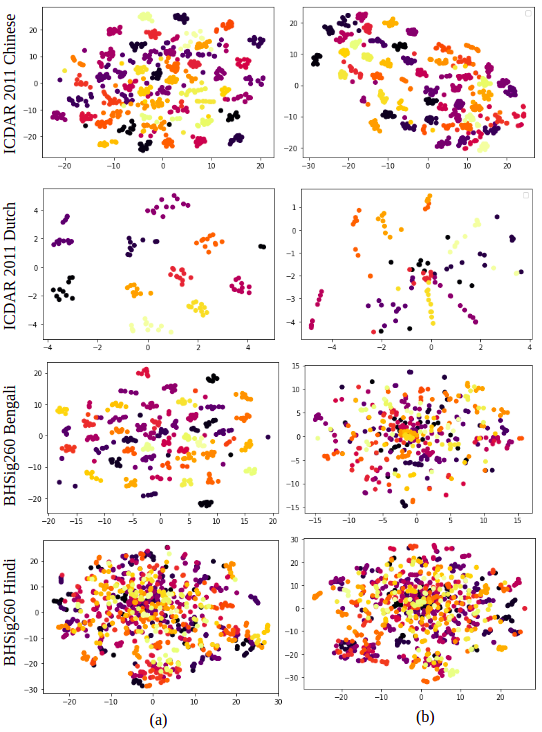}
    \caption{t-SNE visualisations obtained by (a) the proposed method compared with those obtained by (b) SimCLR \cite{chen2020simple}  on different datasets. The color coding scheme denotes each writer cluster. \label{fig:comp_tsne}.}
\end{figure}
\begin{table*}[t]
\caption{Comparison of the proposed method with state-of-the-art self-supervised learning baselines. \label{tab:comp_ssl}}
\centering
\resizebox{\textwidth}{!}{
\begin{tabular}{c|ccc|ccc|ccc|ccc} 
\hline
\multirow{2}{*}{\textbf{Method}} 
& \multicolumn{3}{c|}{\textbf{ICDAR 2011 Dutch} \cite{alvarez2016offline}}     & \multicolumn{3}{c}{\textbf{ICDAR 2011 Chinese} \cite{alvarez2016offline}}
& \multicolumn{3}{c|}{\textbf{BHSig260 Bengali} \cite{pal2016performance}}     & \multicolumn{3}{c|}{\textbf{BHSig260 Hindi} \cite{pal2016performance}}
\\ \cline{2-13} & 
\textbf{Accuracy (\%)} & \textbf{FAR} & \textbf{FRR} & 
\textbf{Accuracy (\%)} & \textbf{FAR} & \textbf{FRR} & 
\textbf{Accuracy (\%)} & \textbf{FAR} & \textbf{FRR} & 
\textbf{Accuracy (\%)} & \textbf{FAR} & \textbf{FAR} \\ 
\hline
SimCLR \cite{chen2020simple}                      
& 69.46                         & 0.554                 & 0.060 
& 59.76                         & 0.431                 & \textbf{0.317}
& \textbf{73.45}                & \textbf{0.117}        & 0.543        
& \textbf{72.45}                & \textbf{0.103}        & 0.599           \\
\textbf{Proposed}                 
& \textbf{77.62}                & \textbf{0.316}        & \textbf{0.133}       
& \textbf{64.68}                & \textbf{0.278}        & 0.583
& 72.04                         & 0.367                 & \textbf{0.116}        
& 72.43                         & 0.104                 & \textbf{0.598}  \\
\hline
\end{tabular}
}
\end{table*}

\section{Experimental Details}
\label{sec:exptdet}

In this section, we are going to discuss the details of the datasets that were used in our experiments, and the configurations used for training our encoder in the pretext (or pretraining) task. 

\subsection{Datasets}
\label{subsec:dataset}
In this work, we used two datasets, namely, BHSig260 \cite{pal2016performance} and ICDAR 2011 \cite{alvarez2016offline}. BHSig260 dataset contains signatures from 100 writers for Bengali and 160 writers for Hindi signatures. For each writer of both the languages, there are 24 genuine and 30 forged signatures. Among the 100 writers in the Bengali subset, we randomly select 50 writers for the training set and the rest 50 are used for testing. For the Hindi subset, we randomly selected 50 writers for self-supervised pretraining and the rest 110 writers were left for testing. Similarly, for ICDAR 2011 Signature Verification dataset, there are signatures for Dutch and Chinese languages. The subset of the Dutch signatures contains signatures from 10 writers for training and 54 writers for testing. %The divisions are however given in the dataset. %The Dutch subset contains 362 signatures in the training set and 1932 signatures in the test set. Among the 1932 signatures in the test set, 646 signatures are used as reference signatures for the writers and the rest as the unknown samples. Similarly, for the Chinese subset, the reference or training subset and the test set, both contain signatures from 10 writers. The number of signatures in the training and test set in the Chinese subset is 575 and 602, respectively. Among the 602 signatures in the test set, 115 signatures are used as reference signatures and the rest are used as the unknown samples. 
In the test set, however, there are 8 reference genuine signatures for each writer. To adhere to this structure, we randomly selected 8 genuine signatures from the test set of BHSig260 dataset for each writer and used it as the reference set, for both Bengali and Hindi languages.

%GPDS300

%The train and test set divisions used in our work is described in the table below:

% \begin{table}[tbp]
%     \centering
%     \caption{Details of the signature verification datasets}
%     \label{tab:datadivtable}
%     \begin{tabular}{c|c|c}
%     \hline
%         Dataset & \multicolumn{2}{c}{No. of writers}\\ \cline{2-3}
%         & Training & Test  \\ \hline
%         ICDAR2011 Dutch & 10 & 54\\ \hline
%         ICDAR2011 Chinese & 10 & 10\\ \hline
%         BHSig260 Bengali & 50 & 50 \\ \hline
%         BHSig260 Hindi & 50 & 110 \\ \hline
%         %GPDS300 & 150 & 150 \\ \hline
%     \end{tabular}
% \end{table}

\begin{table}[t]
\caption{Comparison of the proposed method with supervised learning methods in literature. \label{tab:comp_supervised}}
\centering
\resizebox{\linewidth}{!}{
\begin{tabular}{c|ccc|ccc} 
\hline
\multirow{2}{*}{\textbf{Method}} 
& \multicolumn{3}{c|}{\textbf{BHSig260 Bengali} \cite{pal2016performance}}     & \multicolumn{3}{c}{\textbf{BHSig260 Hindi} \cite{pal2016performance}}
\\ \cline{2-7} & 
\textbf{Accuracy (\%)} & \textbf{FAR}   & \textbf{FRR}   & 
\textbf{Accuracy (\%)} & \textbf{FAR}   & \textbf{FRR}     \\ 
\hline
Pal et al. \cite{pal2016performance}   
& 66.18                  & 0.3382         & 0.3382          
& 75.53                  & 0.2447         & 0.2447
\\
Dutta et al. \cite{dutta2016compact}
& 84.90                  & 0.1578         & 0.1443         
& 85.90                  & 0.1310         & 0.1509 
\\
Dey et al. \cite{dey2017signet}                           
& 86.11                  & 0.1389         & 0.1389          
& 84.64                  & 0.1536         & 0.1536
\\ 
Alaei et al. \cite{alaei2017efficient}
& --                     & 0.1618         & 0.3012
& --                     & 0.1618         & 0.3012              
\\ \hline
\textbf{Proposed}                 
& \textbf{72.04}         & \textbf{0.367} & \textbf{0.116} 
& \textbf{72.43}         & \textbf{0.104} & \textbf{0.598}

\\
\hline
\end{tabular}
}
\end{table}

\subsection{Pretraining Experiments Configuration}
\label{subsec:preexptconfig}
For the pretraining phase, we used different number of epochs for different datasets. The models were trained by optimizing the loss function given by \ref{eqn:lc} using LARS \cite{you2017large} optimizer. We used a learning rate of 0.1 and a momentum value of 0.9. The batch-normalization and bias parameters were excluded from weight normalization. We decayed the learning rate following a cosine decay schedule with a linear warmup period of 10 epochs at the start. The decay was scheduler for 1000 epochs irrespective of the number of training epochs. 

For the ICDAR datasets, we pretrained the model for 500 epochs. Whereas for the BHSig260 dataset, the pretraining was carried out for 200 epochs only. For both the datasets, the batch size used was 32. 

To ensure that the pretrained models learn generalized and robust features, we applied several augmentations, such as, color jittering, affine transformation and random cropping to $224 \times 224$. The images obtained after augmentation were normalized to the range $[-1.0, +1.0]$. 

As not all images in the datasets contain perfectly cropped signature images, we cropped the images such that the input to the encoder contained is a tightly bounded signature image. To achieve this objective, we performed Otsu's thresholding \cite{otsu1979threshold} followed by finding the bounding box with least area containing all non-zero pixels around the centre of mass of the image. After this preprocessing step, the images were divided into patches of dimension $32 \times 32$ with an overlap of $16$ pixels and fed to the encoder for training.

\section{Experimental Results}
\label{sec:results}
\subsection{Downstream Results}
\label{subsec:downres}

The downstream task we considered in our work is the writer-independent classification of signatures into two classes: genuine or forged. The predictions were obtained using the procedure described in Section \ref{subsec:downmethod}. The results obtained by the proposed model in the downstream task on the datasets ICDAR 2011 and BHSig260 signature verification datasets are given in \autoref{tab:comp_ssl}. %These two datasets contain large number of signatures in different languages to validate the effectiveness of the proposed model. 
\textcolor{black}{We also pre-trained and validated our proposed method on GPDS300 \cite{gpds300} and CEDAR \cite{cedar} dataset, and we achieved accuracies of $69.28\%$ and $83.8\%$, respectively. }

\subsection{\textcolor{black}{Ablation on Hyperparameters}}
\label{subsec:ablhyp}
\textcolor{black}{
 We tested the robustness of the representations learnt by our proposed model using Gaussian noise(AWGN) with $\mu=0.0$, $\sigma^2=0.01$ and obtained accuracy(ACC), FAR and FRR of $76.84\%(\sigma=0.26533)$, $0.3242(\sigma=0.005)$ and $0.17(\sigma=0.003)$, respectively for the CEDAR dataset. Using Random cropping, we  obtained ACC, FAR and FRR of $79.3\%(\sigma=0.94)$, $0.344(\sigma=0.0124)$ and $0.1157(\sigma=0.0128)$, respectively. We also consider ablation on projector depth, augmentation and patch overlap on the CEDAR dataset. Increasing the overlap of patches from 0 to 8 pixels shows accuracy(ACC), FAR and FRR of $83.8\%$, $0.118$ and $0.187$, respectively. Increasing the number of layers in the projector did not improve the performance. Removing color jitter as augmentation from the above model yielded ACC, FAR and FRR of $83.1\%$, $0.11$ and $0.19$, respectively.
}

\subsection{Comparison with SOTA Self-supervised Algorithms}
\label{subsec:comp_ssl_sec}

In this section, we show how the proposed loss function fares at training the encoder to learn representations from the data. As shown in \autoref{tab:comp_ssl}, in spite of trained in a self-supervised manner, the proposed framework performs satisfactorily on both the multilingual datasets. \autoref{tab:comp_ssl} also presents the comparative results of one of the state-of-the-art self-supervised algorithm (SimCLR) on the same data.
From Fig. \ref{fig:comp_tsne}, we can see that the proposed algorithm performs better at producing distinct clusters for ICDAR 2011 Chinese and BHSig260 Bengali dataset, whereas the plots for ICDAR 2011 Dutch and BHSig260 Hindi datasets look equally well-clustered for both the proposed model and SimCLR. It should be mentioned here that the SimCLR algorithm was trained for 1000 epochs on the ICDAR 2011 dataset (both, Dutch and Chinese). %, whereas the proposed model was trained for only 500 epochs on the same datasets. Both the methods were trained for 200 epochs on the BHSig260 dataset. %Thus, we can say that the proposed algorithm has the capability of performing better than SimCLR even with less training \textcolor{red}{Need to rephrase the claim. We have trained for same no. of epochs in case of BHSig.}.

\subsection{Comparison with Supervised Methods}
\label{subsec:comp_sup}

To further validate our proposed self-supervised pipeline, we compare its performance with some fully supervised methods in literature. The results have been tabulated in \autoref{tab:comp_supervised}. We observe that the proposed framework performs competitively against the fully supervised works on the BHSig260 datasets, outperforming \cite{pal2016performance} by a large margin on the Bengali signature dataset. Moreover, the low FAR and FRR values obtained by the proposed method on the signature datasets affirm its potential in separating forged signatures from the genuine ones. %\textcolor{red} {Remove ICDAR from Table II if the values are not available.}.

\section{Conclusion}
\label{sec:concl}

In this work, we proposed a self-supervised representation learning framework where a novel loss function is used that aims at decorrelating the dimensions from each other to discard redundant features and encourage learning of linearly uncorrelated generative features of the input. Through t-SNE plots we show that the proposed algorithm extracts better uncorrelated information from the input than the SOTA SSL methods on the same datasets. From the comparative results, it is evident that the proposed method performs better than or at par with the state-of-the-art algorithm SimCLR. This work shows the extensive scope and applicability of the proposed method in the field of signature verification and paves a way for further research in this direction.

% references section
\bibliographystyle{IEEEtran}
\bibliography{refs}

\end{document}